# GRAPHYP: A SCIENTIFIC KNOWLEDGE GRAPH WITH MANIFOLD SUBNETWORKS OF COMMUNITIES

## Detection of Scholarly Disputes in Adversarial Information Routes

Renaud Fabre, ORCID 0000-0003-4170-324X

*Dionysian Economics Laboratory (LED), Université Paris 8, Saint-Denis, France*

Otmane Azeroual, ORCID 0000-0002-5225-389X

*German Centre for Higher Education Research and Science Studies (DZHW), Berlin, Germany*

Patrice Bellot, ORCID 0000-0001-8698-5055

*Aix Marseille Univ, Université de Toulon, CNRS, LIS, Marseille, France*

Joachim Schöpfel, ORCID 0000-0002-4000-807X

*Univ. Lille, ULR 4073 - GERiiCO - Groupe d'Études et de Recherche Interdisciplinaire en Information et Communication, 59000 Lille, France*

Daniel Egret*, ORCID 0000-0003-1605-7047

*Université Paris Sciences & Lettres, 75006 Paris, France*

*\*Corresponding author: daniel.egret@psl.eu*

### ABSTRACT

The cognitive manifold of published content is currently expanding in all areas of science. However, Scientific Knowledge Graphs (SKGs) only provide poor pictures of the adversarial directions and scientific controversies that feed the production of knowledge. In this Article, we tackle the understanding of the design of the information space of a cognitive representation of research activities, and of related bottlenecks that affect search interfaces, in the mapping of structured objects into graphs. We propose, with SKG GRAPHYP, a novel graph designed geometric architecture which optimizes both the detection of the knowledge manifold of "cognitive communities", and the representation of alternative paths to adversarial answers to a research question, for instance in the context of academic disputes.

With a methodology for designing "Manifold Subnetworks of Cognitive Communities", GRAPHYP provides a classification of distinct search paths in a research field. Users are detected from the variety of their search practices and classified in "Cognitive communities" from the analysis of the search history of their logs of scientific documentation. The manifold of practices is expressed from metrics of differentiated uses by triplets of nodes shaped into symmetrical graph subnetworks, with the following three parameters: Mass, Intensity, and Variety.

## 1. Introduction: Representing the manifold of cognitive communities

To address the challenges of research impact assessment, new technical solutions are becoming increasingly popular for representing scholarly knowledge through scientific knowledge graphs (SKGs) (Manghi et al., 2021). SKGs create interactions of nodes that explore information spaces representing research results. Unfortunately, SKGs suffer from the flaw of being barely designed to compare paths leading to contradictory results, a situation frequently encountered in scientific approaches, and more specifically in the context of scientific controversies. This bottleneck severely limits the expressive power of the SKGs in representing the research workflow in its entirety and versatility as, indeed, the results of a documentary query may involve quite different methods, assumptions, or theoretical approaches and conditions. The ability to architect heterogeneous means of information retrieval is identified as a key to the completeness of the knowledge graph.

This article will explore emerging SKGs technologies leveraging the variety of possible answers to a research question: capturing this rich diversity opens new avenues and tools to identify scientific and documentary approaches developed in specific "cognitive communities". The term "cognitive communities" is commonly used in studies relating to SKGs to designate groups of users sharing a common profile in their search for information. These communities are delineated as entities sharing similar pathways to answers, revealed by data captured on their documentary profile and search history on a given research question. Data capture is very efficient because meaningful structures partitioned into communities reveal strongly connected groups of online documents (Rossetti & Cazabet, 2019).

The expression of the manifold of "cognitive communities", which scholarly dispute activates, has been traditional in academia, since medieval times, because it strongly contributes to the advancement of knowledge through the confrontation of contradictory opinions, and the systematic refutation of opposing statements. However, while this immemorial practice provides invaluable traceability to knowledge pathways leading to scientific results, it seems paradoxical in the digital age that the expression of scholarly disputes is barely achievable with current SKG architectures. Indeed, data structures that would identify the expressive power of adversarial scientific communities with node embedding techniques, are not commonly available today[1]. The search for a solution to the problem of representing and delimiting alternative scientific paths and outcomes is addressed in many recent publications, with the aim to alleviate obstacles to recall and integrity in the assessment of research impact. The common objective is to make visible in the knowledge graphs, not only the scientific result itself, but the complete path towards new knowledge, and its place among other alternative paths. We must observe here that there are limits to the discoverability of alternatives to new knowledge from "cognitive communities" in all related fields, which does not yield better results: on a more global scale, practices of knowledge recombination, in the digital turn, still lack a comprehensive understanding of their intrinsic nature (Hund et al., 2021).

Bottlenecks in the design of SKGs have well known consequences, as they prevent research impact assessment from delineating the real-world behaviors of the challenging « invisible colleges » (Zitt et al., 2019) in scholarly publication. Nevertheless, the need exists and recording user experience in a diversity of knowledge expressions is a well-known claim: the Open Knowledge Research Graph project (Jaradeh et al., 2019) recommends the use of « techniques that acquire scholarly knowledge in machine actionable form as knowledge is generated along the research lifecycle ». This goal is considered a hot priority: « Organizing scholarly knowledge is one of the most pressing tasks for solving current and upcoming societal challenges » (Jaradeh et al., 2021).

## 1.1 Research question

The gap observed between identified needs and available technology suggests the magnitude of the methodological challenges. This research question is the subject of many recent developments that we will attempt to briefly discuss below.

On one side, the endemic growth of the cognitive manifold of published content is a feature of contemporary knowledge construction practices, as can be seen from a variety of metrics: observed differences in initial opinions towards the efficient advancement of science have been extensively studied with proposed new metrics of « atypicality » and « disruption » in the assessment of scientific publication; these new indices testify to the positive weight of « disconnection » and « discord » (Lin, Evans, & Wu, 2020) on the construction and advancement of fruitful orientations in scientific ideas and projects. Moreover, the interactions between categories of scientific vocabularies, measured by metrics of the « cognitive extent of science » (Milojević, 2015), denote greater interdisciplinarity with larger networks of collaborations, in most disciplines.

---

[1] We will show in Section 2 how recent literature has taken up the new concept of "generative adversarial networks" in the context of machine learning.

On the other side, a recent comprehensive analysis of trends in SKG outcomes (Manghi et al., 2021) observes that SKG efficiency is primarily significant in global encyclopedic databases of scientific results, across « classic metadata ». However, Knowledge Graph "incompleteness" causes serious obstacles to the expressive power of graphs as a whole (Hogan et al., 2021)[2], and even more in the applications of Scientific Knowledge Graphs (Nayyeri et al., 2021) and namely in full research activities, where SKGs perform poorly (Open Research Knowledge Graph[3], MAG…), « due to the heterogeneity, inhomogeneity, and evolution of scholarly communication » (Jaradeh et al., 2021) while it is clear that « these KGs are ambiguous due to a lack of standard terminology used across the literature and poses domain-specific challenges for KG completion task ». Trade-offs could be identified between *sparsity reduction* and *complexity admittance* in representation of scientific results, which are « incomplete by design » as science explores always a supposedly missing path (Destandeau & Fekete, 2021). Moreover, existing solutions delivered by the SKGs « are still relatively static » (Chen et al., 2020).

Considering these challenges to the contributions of SKGs to the assessment of research impact, our study addresses the issue of understanding bottlenecks in knowledge representations, in search interfaces, and in matching structured objects in graphs applied to cognitive information. As a scientific baseline, we refer, as the application of C.E. Shannon's theory remarks, to the idea that « a computational scheme of a cognitive process, may itself be deemed as a form of cognition ». In that direction, we note also that a "cognitive information space" has been mathematically described and is shaped with differentiable manifolds, representing as proposed, a « manifold-atlas topology » of cognitive systems (Glazebrook & Wallace, 2009). Our study aims to obtain new results in the design and realization of SKGs, which would make it possible to represent varieties of knowledge of cognitive communities, and which could trace expressive routes of corresponding disputes in academia, and thus map the adversarial answers to a given research question. For this, this article will mainly follow SKG representation approaches combining topological connections and interpretable vectors that participate in the program of Geometrically Equivariant Graph Neural Networks (Han et al., 2022).

## 1.2    Main findings

Our findings include the design and first test of SKG GRAPHYP: instead of delivering a single AI-processed answer to a research question, SKG GRAPHYP represents the landscape of adversarial answers, which a query on a research question might reveal. For that purpose, GRAPHYP adopts new approaches in output representation and interpretability, with a novel methodology for designing "Manifold Subnetworks of Cognitive Communities" (MSCC). GRAPHYP captures documentary selections from SKG users, allowing classification of their search paths in a given research field. Users are detected from the manifold of their search practices and classified in "cognitive communities" from the search history of their logs of scientific documentation. Manifold of practices is expressed from metrics of differentiated uses of documentary resources by triplets of nodes shaped in graph subnetworks, with three

---

parameters: Mass, Intensity, and Variety. Subnetworks of documentary practices are themselves structured in flows of data geometrically oriented by an entity alignment.

The solution that we propose is a message passing crown hypergraph, shaped in symmetric bi-partite equivariant interpretable community networks. As MSCC only represents documentary choices, and is thus neutral to underlying scientific choices, SKG GRAPHYP enables meta-learning analysis, browsing and interpretation, as part of a useful user experience of search, on the multiple pathways from search activities to research results. GRAPHYP thus exploits a new stage in the interpretability of SKGs, with manifold detection of documentary routes, and navigation among the controversies documenting the identified cognitive communities.

The rest of this paper is organized as follows: **Section 2** reviews *Background and Other works*. **Section 3** analyzes GRAPHYP's Manifold subnetworks of Cognitive Communities and Graph Matching representations. **Section 4** presents the summary and perspective of our research.

The flowchart of this Article can be represented as follows:

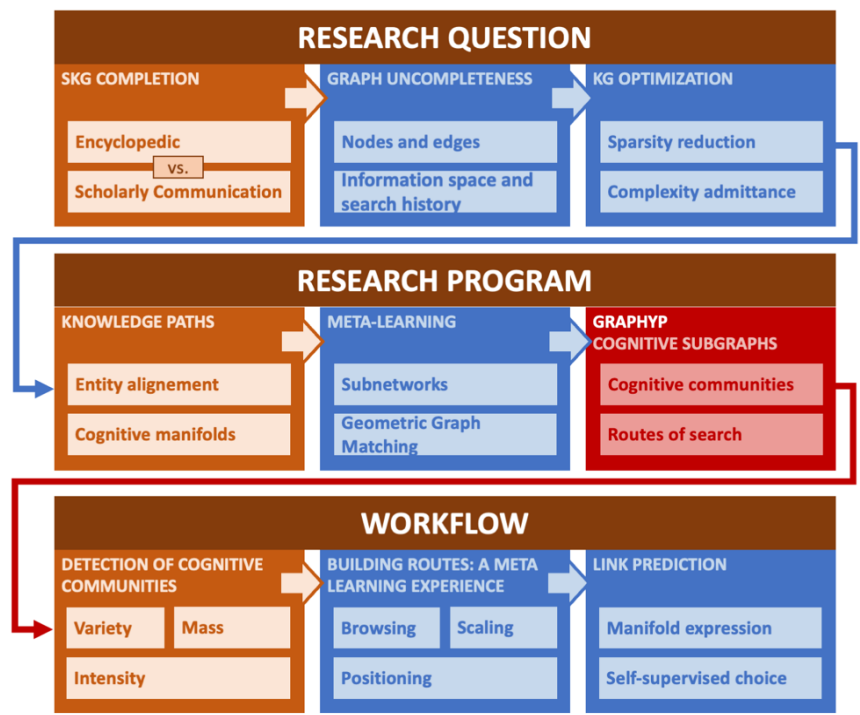

**Figure 1:** GRAPHYP Flowchart: Research Question, Research Program, Workflow.

## 2. Background and other works

Methodologies for SKG design are now confronted to a simple scientific context: digital analytics of scholarly content progresses slowly and is reported as being in an early phase only[4], while the disconnection between research on information behaviors and research on information systems development has been recently highlighted in a comprehensive review (Huvila et al., 2021). Research resources are available, however, in adjacent scientific spheres, and address the considerable structural diversity of digital community networks (Easley & Kleinberg, 2010). A wide range of suitable technologies are available, to express the adversarial learnings of academic dispute, with entity alignment techniques (Munne & Ichise, 2022), with dynamic community discovery (Rossetti & Cazabet, 2019), and with architectures of message-passing graph subnetworks (Bevilacqua et al., 2022)[5] that express the magnitude of manifolds (Chakraborty et al., 2022). User knowledge in informational search sessions has been modeled for the task of predicting knowledge gain and knowledge state of users in Web search sessions (Yu et al., 2021) with limitations due to the limited availability of search session data.

---

[4] See, e.g., the Open Research Knowledge Graph wiki: https://gitlab.com/TIBHannover/orkg/orkg-frontend/-/wikis/home

## 2.1 Background: information space of cognitive communities

The approach to the information space of cognitive systems applied in SKG GRAPHYP is influenced both by symmetry in information, driven by new concepts of differential geometry in graphs (Bronstein et al., 2021), and by ubiquity in networks, which is carried by descriptive applications of graphlets in multiplex (Dimitrova, Petrovski & Kocarev, 2020).

Within this general scientific framework, the background of SKG GRAPHYP with its methodology for designing "*Manifold Subnetworks of Cognitive Communities*" (MSCC), is described below in its scientific environment, while following the successive stages of its operation: (a) cognitive community detection, (b) manifold expression in subnetworks of information, (c) graph matching in a hypergraph of subnetworks, (d) meta-learning experience of the users of SKGs.

### (a) Cognitive community detection

A first step in MSCC is to detect "cognitive communities" from the search history of logs on scientific documentation about a research question: we build digital structures, communities, that ingest the raw material of documentary resources, and we class those communities in dedicated subnetworks of nodes, according to their practices. Technologies are now available for discovery and classification of entities in a wide range of dynamic complex structured networks of nodes, representing communities of practice in the SKGs (Rossetti & Cazabet, 2019) within dedicated subnetworks. In GRAPHYP modeling, the identity and content of communities are processed from the "raw material" of the analysis of search history queries.

Cognitive community detection from query analysis and data from search sessions can be captured from multi-source documents, through the extraction of knowledge instances such as entity, relation, attribute (Chen et al., 2020) with the SKG functioning as a semantic network. To exploit the raw material of queries, complementary logs (Agosti, Crivellari & Di Nunzio, 2009) are processed for an "explicit" semantic ranking via knowledge graph embedding. For instance, a comprehensive study of log analysis of researchers to Semantic Scholar's search Engine (Xiong, Power, & Callan, 2017) has been recently achieved. Applications are developed on Semantic Scholar Academic Graph with an easy-to-use JSON Archive[6]. Transition from raw material to structured communities requires a machinery and steps. The first step is to find original cognitive processes in various contexts of information retrieval, in directions that highlight what could be called « networks of cognitive relevance ». How to express the multiplicity of cognitive practices from metrics of cognitive relevance of the differentiated uses of documentary resources? In SKG GRAPHYP we will use triplets of nodes shaped into graph subnetworks, with three parameters, Mass, Intensity, and Variety which measure the documentary behaviors of the users on the items (see below, 3.2 Subnetworks of Cognitive Communities).

---

Community detection also assumes a data quality assessment approach among the alternatives, and it is proposed to use a classifier as a judge (Ghojogh et al., 2021). New designs in that direction can be mentioned with Generative Adversarial Networks (Vanilla GAN) and Adversarial Autoencoders.

### (b) Manifold of the Cognitive Communities featured in subnetworks

A second step in the methodology of MSCC involves finding a modeling of manifolds (of the search results) in the form of structured subnetworks.

Manifold learning is currently attracting attention (Roweis & Saul, 2000) and has recently become a popular field for its ability to propose efficient solutions in subdivisions of networks in a graph: « Think globally, fit locally » as expressed by the associated slogan. Manifold learning currently refers to the user's specific experience in various real-world practices (e.g., biomedicine, radiology), the information of which is embedded into alternative subnetworks, between which the user must select one, or even a network of subnetworks. It is observed that the modeling of the manifold representations of subnetworks actualizes the ideas of C.E. Shannon on symmetry in information: actualization here means describing information with systems of cognitive variety, whose representations are structured with symmetry, as analyzed below.

A new family of manifold mapping updates representations in a context where knowledge variety has been defined as a « causal pathway » to information, requiring its own descriptive geometry. The authors of the concept of causal pathway, Glazebrook & Wallace (2009), coming from the field of differential geometry, base their approach on the assumption of the *endemic diversity* of the fundamental concepts of information: « any information space could be viewed as in part representing a 'causal pathway' embedded within some culture, but included are semantic, dynamic principles seeking to incorporate states of experience, properties often lacking in traditional cognitive theories ».

In such an approach, Graphs are supposed to have their own added value in representing the cognitive manifold of information, with the application of a specific expressive symmetry in the construction of information. Glazebrook & Wallace (2009), also referring to C.E. Shannon with the concept of "Rate Distortion Manifolds" and originating with « necessary conditions gauging the reliability of a source entropy rate relative to a channel capacity », are defining a research program for information representation within which our own research question is housed with SKG GRAPHYP representation of manifolds. In their research program, the authors observe: « introducing simplicial methods to analyze the underlying combinatorial structure of the manifold, we may recover graph-theoretic models as suited to the navigation through various types of information highways, systems of coding, symbolic dynamics and complexity». Their reference to Shannon suggests that they foresee an extension of the well-known theoretical condition of symmetry to a representation of "symmetrized manifolds", and thus that Shannon's theory could find an extension to a represented symmetry of (cognitive) information systems. This powerful concept is at work today, in approaches such as the use of hypergraph manifold regularization, to maintain consistent relationships between the original data, transformed data, and soft labels (Shao et al., 2022).

### (c) Consistent manifold of search history of Cognitive Communities

In our MSCC methodology, representing cognitive manifolds requires suitable homogeneous data. SKG users' search history is a rich data stream to build consistent cognitive manifolds and drive relevance in the SKGs. As observed by Eickhoff (2014): "Automatic inference schemes based on document content and user activity can be used to estimate such constituents of relevance ». The parsing of documentary logs on a research question helps to build « semantic paths » to knowledge (Destandau & Fekete, 2021), as coined also by the first comprehensive survey dedicated to Cognitive Graph applications (Chen et al., 2021). User data and user-generated content are thus an endemic component for shaping the « naturally created community information on content sharing platforms to infer potential tags and indexing terms », with the aim «to mitigate the vocabulary gap between content and query » (Eickhoff, 2014).

A large corpus of knowledge is available on retrieval methodology in user interaction frameworks[7] and the breadth of manifold definition has been demonstrated in networks of manifold-valued data (Chakraborty et al., 2022).

### (d) Graph matchings of Subnetworks of Cognitive Communities

Subnetworks of documentary practices are themselves structured in flows of data geometrically oriented by an entity alignment which could make it possible to detect, in a controversy, a preferred community identified in a self-supervised mapping of adversarial representation of all observable communities.

A novel functionality is required to build up the *semantic adversarial representation of cognitive communities of knowledge.* Proposed in the SKG GRAPHYP, this functionality is necessary to identify, within a manifold of communities, typical *adversarial positions and controversies.* This implies first to represent each community in a specific modeling of all "possible" manifolds, within a dedicated subnetwork. Tailored subnetworks are now accessible in Graph Neural Networks (GNN) with a topologically aware message-passing scheme based on substructure encoding (Bouritsas et al., 2021), which successfully overcomes former inability of GNN to detect and count graph substructures, via subgraph isomorphism counting. This illustrates one among many benefits of geometric deep learning models (Bronstein et al., 2021).

The categorization of transactions between communities within a global modeling proceeds from representations of bi-partite graphs: they allow matchings of graphs in two homogeneous projections, with vertices that can be divided into two independent disjoint sets. We can thus build a transaction data base between bi-partite projections in meta-paths that are consistent to triplets of nodes connected from both parts. Graphs are thus modeled as pairs of transactions: each pair formulation has a weighted projection. This technology, proposed by Bruss et al. (2019), involves a procedure that we follow in GRAPHYP with a design specific to the geometry of bipartite graphs (see Section 3.2).

### (e) Meta learning experience in the analysis of documentary routes

A cornerstone concept in this context is that of clustering and embedding of neural manifolds (Tripuraneni, Jin, & Jordan, 2022) recently developed as a meaningful and interpretable feature space, adding a geometric constraint to make the clusters identifiable.

Meta Learning in SKGs could thus take the future paths of « Cognitive Graphs » (Chen et al., 2021). The analysis of sense-making in search activities develops in directions that highlight what might be called « networks of cognitive relevance ». This is explicitly the case with the attractive idea of « cognitive orbits » studied from man-machine interactions in a low-dimensional cognitive manifold and leading to « cognitive topological structures » (Cheng, Liu, & Meng, 2015). A neighbor approach lies in the idea of learning interaction kernels for agent systems on Riemannian manifolds (Maggioni et al., 2021): networks have something to say and might make sense in orienting alternatives in cognitive processes.

---

[7] See, e.g., Proceedings of the Workshop on Understanding the User - Logging and Interpreting User Interactions in Information Search and Retrieval: https://ir.webis.de/anthology/volumes/2009.sigirconf_workshop-2009uiir/

At last, « adaptative interfaces » of the same family as GRAPHYP –knowledge learning from each other– are being developed today in the field of human–machine interactions: these interfaces propose seamless interactions with users during online operations, with closed-loop adaptation of the interface, driven by the user's known movement intention (Rizzoglio et al., 2021).

## 2.2    Other Works on information space of cognitive systems: « learning from predecessors »

Pre-trained language models currently receive much attention, while pre-existing models are considered questionable (Metzler et al., 2021). We have previously mentioned the studies of the « cognitive information space » analyzing a particular cognitive situation (Glazebrook & Wallace, 2009). Learning from information retrieval of predecessors has been widely studied for a long time in the fields of web search procedures. However, the need of a global modeling in an ecosystem of self-supervised approaches (Fabre 2019) persists and could fruitfully benefit from adjacent works[8] in fields where *indirect communication mediated by modifications of the environment* is intensively studied. The interdisciplinary field on "learning from predecessors" has developed rich content applied to graph heuristics, within which we position our future work on hypertexts and graphs: this includes the multifaceted study of texts in philology of graphs (Weber, 2020), social stigmergic cognition (Marsh & Onof, 2008), or even the stimulating idea of knowledge graphs as rhizomes in the sense of G. Deleuze[9], as well as new philosophical approaches to the hyperlinking of texts, structured from an integrated "Topology of mind" (Logan & Pruska-Oldenhof, 2022).

**3. Approach.** SKG GRAPHYP: Detection and Search of Cognitive Communities in Geometric Adversarial Information Routes

The design and functionalities of the SKG GRAPHYP inherit the information space characteristics of the cognitive systems discussed in Section 2. These systems can be represented as manifold subnetworks of cognitive communities, which a meta learning experience can detect and analyze, allowing to identify the adversarial ways that could be followed to answer a research question.

In Section 3 we apply this background context to the design of SKG GRAPHYP, for the detection and search of cognitive communities by modeling Adversarial Geometric Information Routes. The objective is to analyze and assess knowledge construction operations as revealed by the search history. We observe that processes of knowledge building, captured by the search history, raise specific modelable characteristics, delineating and capturing different ways of obtaining research results[10] among which *science has to choose and to optimize its access routes to results*. In this field, research impact assessment is challenging in that it must establish a relevant demarcation between the directions traced by scientific outcomes and their routes, and identify the associated semantic changes in the scientific vocabulary, as well as the related practices in data stewardship. These tasks are included in the purpose of the SKG GRAPHYP.

---

[10]The versions of the construction of knowledge depend on the distributed relations of knowledge and their contexts of use; however, for the same context, there may be various hypotheses on the interpretations of the facts and therefore on the possible outcomes. The experimentation itself can follow varied protocols according to varying hypotheses, and thus the same initial knowledge leads either to alternative ways to reach alternative outcomes, or to the same outcomes reached by alternative ways. *Comparing ways to outcomes is one source of research assessment and one strategic way to exploit search history in GRAPHYP.*

Thus, in this Section, our raw material is the search history, and the units of analysis are the cognitive communities as nodes, relevantly linked in a SKG on alternative routes. We will first present the general design of SKG GRAPHYP information space, as a cognitive system of communities of search (purpose, functions and modeling); Subsection 3.2 reviews detection and integration of subnetworks of cognitive communities in a hypergraph; Subsection 3.3 presents our first tests on triplet adjustment. Finally, we discuss the subnetwork analysis and the comparisons of cognitive community practices (Subsection 3.4).

### 3.1    Design of the SKG GRAPHYP: building the information space of a cognitive system

<u>(i) General modeling and problem set-up</u>

<u>Purpose</u>: The purpose of GRAPHYP is to represent the plurality of research answers that can be categorized from the captured diversity of search histories on a scientific subject, thus representing the adversarial versions of the answers given to a research question.

<u>Method:</u> We follow recent approaches to scientific and technological information oriented "semantics-adversarial" and "media-adversarial" using cross-media retrieval methods, to find "effective subspaces" (Li et al., 2022).

<u>Originality:</u> While classic SKGs provide an answer at the end, the purpose of GRAPHYP is to represent plural answers with categorized manifolds of search histories on a scientific topic. Our approach is close to the recently featured category of systems of « model-based information retrieval », which articulate indexing, retrieval, and ranking, built from a single corpus: GRAPHYP is close to that multipurpose group of multi-task learners for multiple information retrieval tasks (Metzler et al., 2021).

<u>Output and uses:</u> the purpose of GRAPHYP is to represent "Manifold Subnetworks of Cognitive Communities" corresponding to the type of SKG described in section 2 as Generative Adversarial Networks (GAN; Liu et al., 2019). It models antagonic subgraph features and articulates underlying topological distribution of graph structures. This modeling is proposed at different scales and levels of granularity in a "generative" architecture, adjustable to needs, with an improved training ability (Liu et al., 2021). It can thus be used to discover new graph structures and to generate evolving graphs.

<u>(ii) The Design of SKG GRAPHYP</u>

The graph design of SKG GRAPHYP is a crown bi-partite graph with connected edges[11] featured with distance-transitivity[12]. A first sketch of this modeling has already been described in (Fabre, 2019). It is developed and enriched in the Annex to this article, with an example and a methodology for graph matching.
SKG GRAPHYP achieves its purpose by positioning search communities in a "searchable space" where all user communities share the same keywords. The geometry of GRAPHYP allows two functions:
     - it allows each community to be positioned in the searchable space, according to the characteristics of its search history;
     - it assists a community in navigating on the graph, to reach the position of neighboring communities, linked by the same characteristics of search goals[13].

While linking facets of query choice into distinct clusters of similar characteristics, or cognitive communities, GRAPHYP modeling formalizes a "searchable range" of alternative preferences that can

---

be selected from the same search into "communities" of identified users. With an original positioning of the user's current selection in its entire context, this modeling makes it possible to detect "routes" formerly unknown to the user and to compare these routes.

GRAPHYP thus provides a service linking **data** (articles for example) **and their attributes** (readings for example) and represents communities of search recorded at the global scale of a keyword. To our knowledge, no service of this kind is yet open to consultation or barely sketched (Fabre, 2019) for applications to alternative documentary selections or "search profiles", and thus « it remains difficult to identify roots of controversies: accurate interpretation of views, downloads, citations, is uncertain » while « searching can sometimes appear as a lonely walk in a forest of hazy homonyms. » (Fabre, 2019).

## 3.2    Subnetworks of Cognitive Communities: detection and integration in the SKG GRAPHYP

The basic subnetwork unit in the SKG GRAPHYP must be relevant, simple, and duplicable, to adapt to the constraints of MSCC modeling. As shown in Section 2, GRAPHYP's subnetworks are in line with equivariant subgraph aggregation networks recently described and perform well on multiple graph classification benchmarks (Bevilacqua et al., 2022).

We will now examine (i) Cognitive Community definition and detection and then (ii) Entity alignment. The weighting of graph subnetwork matches is discussed in the Annex.

 (i) Cognitive Community: definition and detection

- **Positioning of Cognitive Communities in the SKG geometry of a Bipartite Crown Hypergraph**
User preferences are expressed by choices that the search history records in a modeled matrix of possible choices. This modeled matrix is designed by a mini-max system of topological locations of Cognitive Communities, between two extreme preferences within the framework of GRAPHYP. Between these extrema, recording typical search sessions helps to model the full range of information-seeking needs of user communities for similarly formulated search sessions. GRAPHYP models all the possible choices made from the same question. Our graph construction borrows the unfrequented methodology of a "map equation" formulating a data flow network (Rosvall, Axelsson, & Bergstrom, 2010; first described in Rosvall & Bergstrom, 2008): it infers that programmed « links in a network induce movement across the network and result in system-wide interdependence ». Any Cognitive Community will be defined hereunder by co-integration in the SKG of triplets of values of the parameters recording the search history, with the metrics of Mass, Intensity, and Variety.

- **Basic identification of a Cognitive Community on its Search Route: Mass and Intensity of Nodes in Search History**

Let us develop in operational terms the methodology initially sketched in (Fabre 2019) which has been transformed here in terms of calculation bases. In order to record the "search routes" of a community for a given keyword, routes that may differ and are intended to be compared, we write:

$$Q_n = f(N_n; K_n)$$

where $Q_n$ is the number of searches related to a given topic. Altogether, Q expresses the quantitative weight of any community as measured by the documentary usages of this community for its cognitive purpose of answering a research question. Also, Q could compile different queries provided they belong to the same perimeter of research question.

Let us also consider that for each search Q, we identify a parameter of **Mass** which corresponds to a number N of users, and a parameter of **Intensity** K which corresponds to the number of items (URL, documents, articles) constituting the search outcomes, among a corpus of items related to the keyword or the group of related keywords. We can consider the N users of K items as a community of users of the same query route $Q^{14}$. The positioning of any distinct route can be expressed for a given search within the limits of a system of typical search sessions.

To fix ideas, consider the two limits between which search sessions will fall: the first "limit community " is one in which a few users request many items, and the other limit community is one where many users consult a very small number of items. We can write these possible substitution limits between users and items, N and K:

$$Q1 = f(Nmax, Kmin) \quad or \quad Q2 = f(Nmin, Kmax)$$

Or, in a general form:

$$Q_n = \text{fN}, K \quad \begin{matrix} [\text{max, min} \\ \\ [\text{max, min} \end{matrix}$$

We observe that the search function Q described above, coupling N users and K items, can be described in the same way, if for any query Q, one can record pairs of variables expressing different types of search preferences and choices. More generally the function can couple any selection of items N and items K which takes place within the perimeter of a search.

- Recording dynamics of Search sessions: a third Node measuring the value of a parameter of Variety

Here we propose a new method of recording the dynamics of GRAPHYP which clarifies and differs from (Fabre, 2019). Let us calculate the mean values of N and K on the whole set of search sessions: we can normalize the presentation of all search sessions, as located above or below the mean ratio N/K. An additional information on that ratio is given by its dynamics at the scale of the whole set of analyzed search sessions. In fact, with any recorded value of the N/K ratio, that we call a ratio of **Variety,** there is an associated index of dynamics which expresses that N/K preferences are recorded either from an abruptly changing behavior or from a steadily increasing or decreasing behavior in reading articles (in our example).

If we decide to represent N/K choices with this additional element of stability/instability of the function, we change our function Q of two variables N, K, into a triplet where the third term linking N and K represents the expression of stability/instability of behaviors. For instance, we could practice community detection of the readers of a usual group of chemistry articles "before" and "after" the publication of a new important article and we would thus notice if this additional publication accelerated or not the readings in peripheral related domains.

Let us measure the value of that third term by a ratio calculated from a value of normalization, expressed from the mean value of N/K. We can consider a fraction $\alpha/\beta$ where $\alpha$ is the numerator calculated from

---

N mean value and β is the denominator derived from the mean value of K. This fraction α/β will vary, consequently, with any recorded group of reader and article values[15]. The value of this fraction brings a specific element of dynamic analysis: it expresses the **Variety** of the Cognitive Community's documentary practice and makes it possible to measure the "stability" or "instability" of the content of a search session, when the quantities N and K increase or decrease between two searches in the same "route" of searches, or between all routes of search, when N and K values are computed on the whole set of a group of searches for the purpose of detecting communities.

The data structure of a search can now be expressed by an observable and recordable index of change in the relation between the number of users and the number of items. We can thus write:

$$Q_n = f \; [\text{nmax,} \overset{\alpha}{\text{min}} \; \text{kmax,} \overset{\beta}{\text{min}}$$

Let us note that the values "min" or "max" of *N* and *K* provide information on the measured quantity of these variables to "produce" Q, but, as observed in (Fabre, 2019) "from one unit of search Q to the other (from one query to the other to the other, for instance), the coefficient of increase or of decrease between value "min" or "max" of *N* and *K* provides an additional index of assessment on the quantity of these variables in the "production" of a search Q."

The fraction α/β provides a dynamic index of the variations recorded in the practices and controversies of scientific communities, revealing quantitatively how "strong" or "weak" they could be, as measured by the variation of the recorded flow of documentary practices: by this approach towards sensitiveness to frequency, it may help detecting differences between communities approaching the same concept by homonyms[16].

<u>(ii) Entity Alignment of Cognitive Communities in GRAPHYP modeling</u>

- **Networking search sessions and Detection of Cognitive Communities in the SKG**

We know that, with the added mix of user and items that it measures, the slope of a set of searches tends to be stable when (α/β) is inferior or equal to 1, as this fraction is tending to its mean value on the whole set of recorded search sessions. The method is here related to the Graph assortativity approaches described in *Wolfram Assortativity*[17].

Conversely, there is also an alternative situation in which (α/β) tends towards + or - infinity, indicating an unstable variation of (α/β), which will increase or decrease sharply: the relations between users and items differ in that case from the average measured by the mean ratio (α/β). This may indicate that the user and item pairs are changing significantly and that a threshold has been reached in the effectiveness of the search process.

In this case, strategically, why continue to "allocate" (n; k) to Q if (α /β) tends toward infinity? This would mean that the combination (n; k) becomes « outperforming", which is questionable when the time comes to decide on a new search Qn+1 with the same characteristics of N and K. It could, for instance, be more interesting to change N or K than to reproduce the same quantities in the next search. More generally, for transition between Qn-1 and Qn, the choice of stability in N and K could be "justified" if we observe that stability prevails with ($\alpha/\beta$) tending to 1 and "questionable" when instability prevails with $\alpha/\beta$ tending to + or- infinity.

---

[15] An alternative procedure could be to note α the coefficient of increase of N and β the coefficient of increase of K when Q varies by one unit when an additional query on the same search is recorded.

[16] For the same new category of items, several communities could be « neutral » to a change of publishing orientation, while others could be reactive.

[17] https://reference.wolfram.com/language/ref/GraphAssortativity.html

Over the course of a long series of search, there will be a *learning process* that will give meaning to the expression of limits in the variations of the proposed investigation. From this perspective, it is possible to set arbitrary limits of variations for a given panel of search sessions, with the "best" and the "worst", for instance:

$$\text{Best} = N_{\max}^{\alpha}, (\boldsymbol{\alpha/\beta}) \to 1, K_{\min}^{\beta}$$

And

$$\text{Worst} = N_{\min}^{\alpha}, (\boldsymbol{\alpha/\beta}) \pm \infty, K_{\max}^{\beta}$$

If we have selected these two opposite limits to search sessions for cognitive communities belonging to a search history, we need to give details on the calculation of our limits.

How to select the minimum and maximum values of N and K? How to identify "best" and "worst" cases? There are several steps to prepare for the calculation of these values:

- **selection of parameters for N and K:**

In (Fabre, 2019), N and K values correspond to the number of users and the number of articles, but it could be any pair of data that can be combined to express the preferences of Cognitive Communities. For instance: number of articles citing the article of Author X and number of articles citing Author Y; or number of readers of article A and number of readers of article B, etc.

- **selection of min and max values of N and K:**

When data on N and K are recorded, their respective <u>minimum</u> and <u>maximum</u> recorded values are identified. The four values N (max & min) and K (max & min) give the quantitative limits of GRAPHYP recording, thus defining the potential location of all intermediate values for a search session. Min and max value of N and K are findable and the return of GRAPHYP will be better when recorded from rather large data series.

- **selection of mean value of coefficients α and β**

The mean fraction is calculated on the whole group of searches: it can provide a value for "positioning" the respective opposite values of this fraction at the scale of the whole group of searches, in association with mean and max values of N and K, and, from this basis, on the whole panel of search sessions between the min and max values of N and K.

- **selection of "best" and "worst" cases for Cognitive Communities**

One must select, from the user's point of view, GRAPHYP results that one considers as "objective" limits of the analysis from the point of view of Cognitive Communities. Those would be the "best" and "worst" situations between which all users of a search session are expected to be positioned. Notice that search sessions can be recorded over time for an individual user, or for a comparative set of communities of users.

When one expects assessing preferences of users of a search session, it is easy to observe if, for instance, Nmin is better than Kmin as the "best" case. In addition, the min and max preferences tested on limits can be modified and the results recorded by GRAPHYP can be tested on other limits, provided that the preferences are formulated with opposite symmetrical choices in order to let GRAPHYP logics work, which supposes having limits for the classification of recorded results.

We can then write that the cognitive communities could express their preferences between:

| "Best" | "Worst" |
|---|---|
| $N_{\max}$ | $N_{\min}$ |
| $(\alpha/\beta) \to 1$ | $(\alpha/\beta) \pm \infty$ |
| $\beta$ | $\beta$ |
| $K_{\min}$ | $K_{\max}$ |

(b) Integration of cognitive communities on search session routes in the SKG GRAPHYP

Based on the above-mentioned expression, our two triplets give shape to a formal graph-based representation of paths between the two limits fixed to the expression of preferences, thus positioning all the non-contradictory solutions existing between these two limits, included within three parameters in triplets. For that, we selected the following bipartite crown graph with connected nodes, which is the only way to give the representation needed by a typical network of nodes.

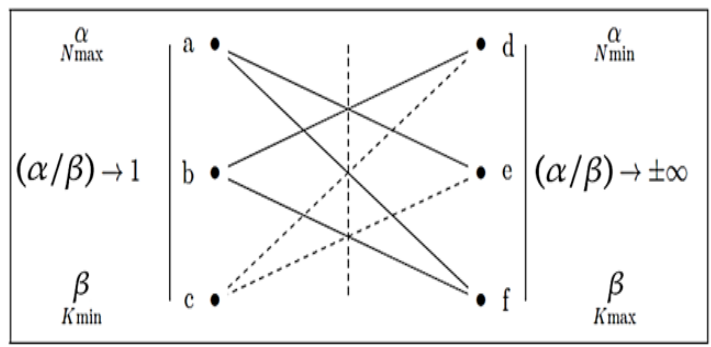

**Figure 2**. "Compass" GRAPHYP: Positioning Typical Search Sessions.

Adapted from (Fabre, 2019)

Figure 2 shows a complete representation of all the typical intermediary situations between our two limits of triplets of nodes, which gives us a tool for the classification of observed search sessions Q in a series of searches on a given keyword, according to the user and item choices. Structured by GRAPHYP, this set of typical search session positions has two main characteristics, which is detailed below with the help of Table 1.

As shown in Table 1, GRAPHYP expresses the whole set of *non-contradictory* positions of combined triplets of nodes, that structured search routes of Cognitive Communities could occupy during the analysis of a group of searches on a given keyword, for instance, on reading articles.

| NODE (Query) | TYPICAL DIRECTION OF SEARCH (Users, Items) |
|---|---|
| a | $N\alpha_{max}, K\beta_{max}, (\alpha/\beta) \to \pm\infty$ |
| b | $N\alpha_{min}, K\beta_{max}, (\alpha/\beta) \to 1$ |
| c | $N\alpha_{min}, K\beta_{min}, (\alpha/\beta) \to \pm\infty$ |
| d | $N\alpha_{min}, K\beta_{min}, (\alpha/\beta) \to 1$ |
| e | $N\alpha_{max}, K\beta_{min}, (\alpha/\beta) \to \pm\infty,$ |
| f | $N\alpha_{max}, K\beta_{max}, (\alpha/\beta) \to 1$ |

**Table 1**. Classification of Typical Search Sessions in GRAPHYP

Adapted from (Fabre, 2019)

For instance, from Figure 2 and Table.1, node *c* is linking edge *cd* and edge *ce*: this node creates the analytic position listed in Table 1, with all typical positions that can be recorded in GRAPHYP data structure, between *a* and *f* positions of search sessions Q: they find their location in this graph as tending to min max values of N, K.

When recording data on a set of search sessions, and preparing them for GRAPHYP uses, we can apply the classifications proposed above to represent, for users, all practices recorded during search sessions issued from the same scientific theme. In section 3.3, we test the conditions to process data on readings in the context of scientific impact assessment.

### 3.3     Tests of GRAPHYP's triplet adjustment

We have successfully tested the robustness and significance of node triplets integrated in the image of Community behaviors. The data come from real-world search history records, in a test panel of our triplets for approximately 10 million search sessions; we are developing in the meantime the analysis of click metrics of scholarly content – a topic currently in full development (Fang et al., 2021).

In order to identify the nature of the data required to implement GRAPHYP in the context of a digital library, we made a first prototype using access log files from OpenEdition.org platforms. These log files were collected by the web analytics platform Matomo[18] and then filtered to eliminate connections from bots and requests for files other than papers. User IDs have been associated to the requests to the Web server according to the anonymized IP address and sessions IDs have been added based on the recorded timestamps. This means that in the following we call session a sequence of actions (content requests) performed by an anonymous user in a limited time.

We can assume that each session thus corresponds to a specific need, even if we do not know the original request made by the reader (the referers haven't been communicating queries for a few years now and the incoming links are usually very vague). In most cases, the reader comes directly from a Web search engine and then follows the links in the pages corresponding to the retrieved papers. We thus have for each session a time-ordered sequence of articles read. Since the queries here are unknown, we can only have the ambition to compare general reader behaviors, regardless of their information needs.

The process is then as follows. First, we divide all the logs into blocks of sessions. Then, for each session, within each block, we estimate the values of K as the number of articles read by a reader. This makes it possible to estimate, in each block, the number of readers N who have read K articles and thus the mean values of the different N and K for this block. $\alpha$ and $\beta$ values as well as their ratio are calculated, for each block of sessions except the first one, from the mean values of N and K.

---

[18] https://matomo.org/about/

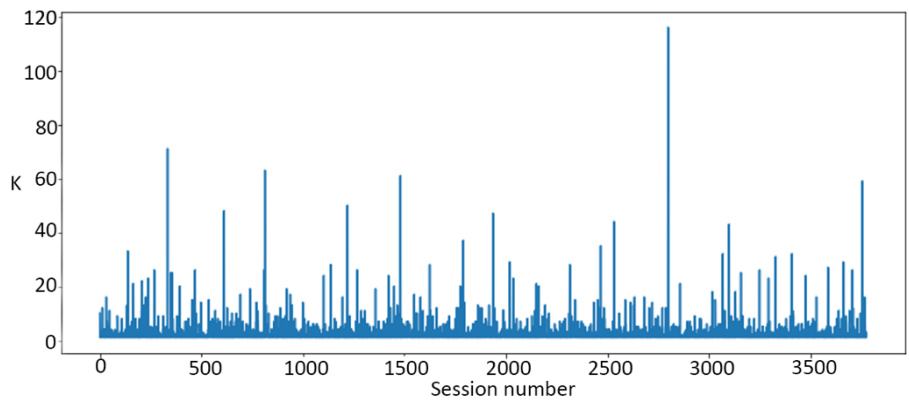

**Figure 3.** K values for about 4000 sessions

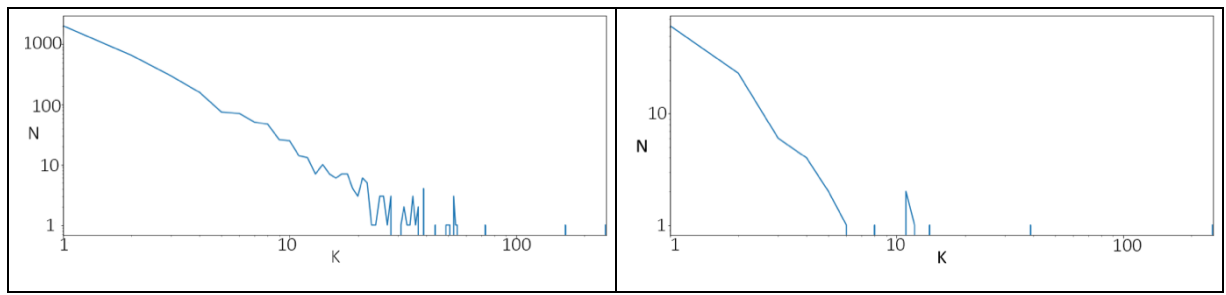

**Figure 4.** N and K values for two different blocks of sessions (on the left, quite a few people read more than 10 articles while on the right most readers read only one article).

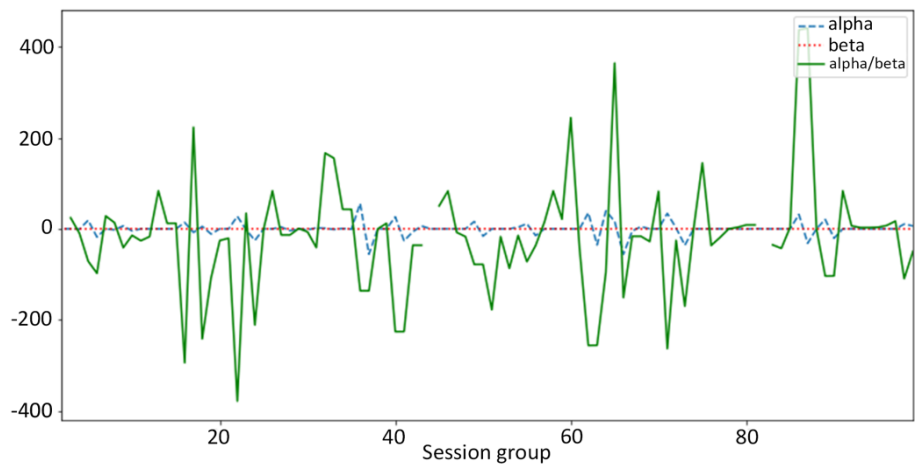

**Figure 5.** $\alpha$, $\beta$ and $\frac{\alpha}{\beta}$ for 100 blocks of 10,000 sessions.

Once this is done, it remains to determine for each session its type according to the associated values of N, K and $\alpha/\beta$. Some thresholds make it possible to identify the tendencies towards the min/max values of N and K and towards ±infinity or stability (1) for $\frac{\alpha}{\beta}$ (for example, we can choose to consider K as type $K_{max}$ if $K > K_{max} - (K_{max} - K_{min}) \times z$ with $0 < z < 1$). Lastly, the sequence of types gives us the routes of search. By considering 100 blocks of 10,000 sessions corresponding to 1 million documents read, 42% of the sessions are of type b. Notably because the blocks are not built according to the user queries (which are not known here), this result was expected: the majority behavior is stable.

**3.4 Subnetwork Analysis and Comparisons of Cognitive Community practices.**

GRAPHYP users can benefit of the three following functions: Scaling, Mapping, Positioning, to determine their interest for a cognitive community, their present distance to it, their estimated route if any, and to measure the distance between the current position and another one.

(i)      Function 1: Positioning a Cognitive Community on a search route
As a sort of "digital compass" GRAPHYP provides all possible "directions of search" and locates the recorded direction of an individual search of Cognitive Communities. It thus implements an original approach to "path-based reasoning" (Jagvaral et al., 2020).

This set of possible "positions" during a search gives an overview of the proposed modeling of the searchable space for informational queries, according to six typical search routes of cognitive communities. Figure 6 shows how GRAPHYP allows to explore typical Neighbor Search sessions, in a context of "betweenness centrality"[19].

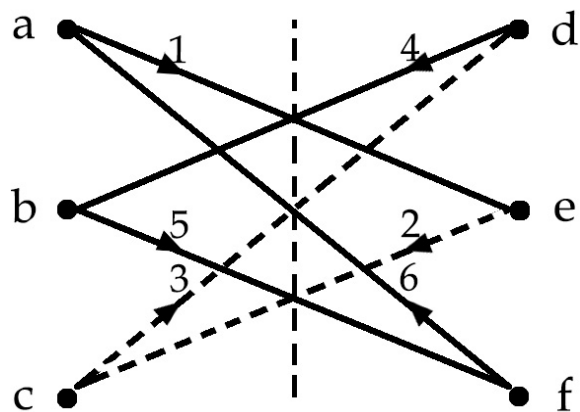

**Figure 6.** Exploring Neighbor Cognitive Communities with GRAPHYP "Compass"
Source: (Fabre, 2019)

Cognitive communities can thus be interconnected and explored, as a continuous circulation from node to node showing a common edge in all pairs of edges corresponding to any node. A circular "travel" is possible within GRAPHYP, from edge to edge, while coming back to the starting node by the way of the complementary edge to that node.

Bipartite graph analysis is an alternative to reveal clusterization in complex systems (Palchykov & Holovatch, 2018), as exploring and analyzing sessions of users could benefit from results produced from the GRAPHYP data structure. We note that research on user-items connections has not currently followed this path and is mostly bound to classical approaches to recommendation applied to SKGs (Wang et al., 2018). Recently Graph Convolutional Networks (GCN) « and variants have achieved state-of-the-art results on classification tasks, especially in semi-supervised learning scenarios. A central challenge in semi-supervised classification consists in how to exploit the maximum of useful information encoded in the unlabeled data » (Pedronette & Latecki, 2021).

The original structure of GRAPHYP thus proposes an *undirected* weighted graph structure (each edge is bi-directional and receives a distinct weight from the nodes linked inside a network), which is connected (one can reach any node from all other nodes inside identified paths) and builds a minimum spanning tree as a subgraph containing all nodes, connected here with the minimum possible number of edges.

---

[19] Betweenness Centrality is a measure of the relative importance of a node based on the number of shortest paths that cross through the node: https://medium.com/rapids-ai/rapids-cugraph-networkx-compatibility-d119e417557c

GRAPHYP gives then a structural approach to least distances solutions in bi-partite graphs representations (it could be borrowed by analyses developed in applications of convolutional networks).

(ii)         <u>Function 2: Mapping positions of Cognitive Communities on search routes</u>

As described in (Fabre, 2019) users of GRAPHYP will be able to learn from their past recorded behavior as well as from viewing the recorded routes of other users of the same base, identifying their "search position" and their "search route": these goals can be reached from a grid of structured data on the recorded searches of all users, as shown in Figure 7.

SKG GRAPHYP may thus be used as a "compass", recording and directing digital navigation in an environment of articles and other scientific resources.

As explained above, the pairs of edges in GRAPHYP always belong to a given node (here $a$, $b$, $c$, $d$….). **Each node designs a grid of proximities between partially neighboring nodes**: we will then exploit those partial proximities to identify paths which, by using next neighbor circulation paths, can:

- allow identifying any optimal "route of preferences" of a user, according to relevant identified proximities of edges and nodes;

- record any observed path of users which, during their past search sessions, have used a recorded way.

Figure 7 shows a genealogy that could be recorded from data on search routes, analyzing the paths of discovery through the assessment of scientific documents characterized by the detection of communities, in laboratories or teams for instance. All distance estimation methods[20] can be applied to this architecture in order to explore neighboring routes, their genealogy, and to map distances and differences between search sessions.

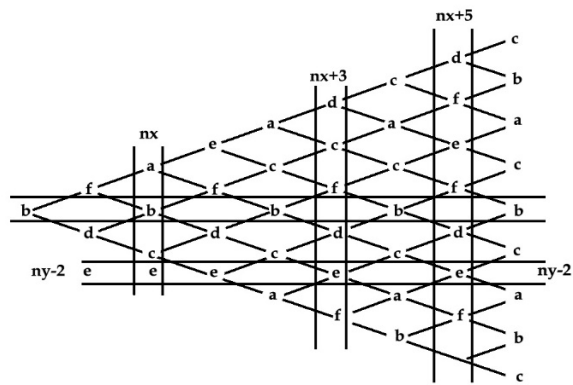

**Figure 7.** Positioning recorded Cognitive Communities on Search History of Search Sessions routes.
Source: Fabre (2019)

In SKG logics, it could be applied to the comparison of discovery paths: documentary choices (readings, downloads, comments, etc.) could provide data for studying the appearance of neologisms and changes in vocabularies. This type of a grid could also allow projections from a given node, onto the closest neighbor positioned nodes and thus provide a basis for a future search strategy.

(iii)         <u>Function 3:  Scaling Cognitive Communities of knowledge extraction</u>

The mapping of informational "searchable routes "of Cognitive Communities is thus designed with rules equivalent to air or sea routes. Cooperative or connecting routes can be identified by the GRAPHYP data structure, which involves adjusting the size of the selected community to an optimized scale. This in turn raises questions of the optimal calibration of subgraphs in the perspective recently called by Michael Bronstein, that of "latent graph learning" and the associated methodology for counting isomorphisms of subgraphs (Bouritsas et al., 2021).

---

[20] https://reference.wolfram.com/language/guide/DistanceAndSimilarityMeasures.html

In this instance, the question for the graph to solve is to develop hyper-relations to reduce features at n-ary scale to smaller tuples. This important research question has recently been successfully clarified by technologies of so-called hyper-relational link prediction (Yan et al., 2022)[21].

Along the same lines, GRAPHYP offers an original and simple structural approach, the intrinsic geometry designed for this SKG being able to be duplicated at any expected scale: as illustrated by Figure 8 below, GRAPHYP with nodes A, B, C, etc. is built by addition of graphs of the same shape, and open thus to a self-similar substitution[22]: this self-similarity characteristic of GRAPHYP (Fabre, 2019) allows knowledge extraction at any scale and allows operating scalability from perimeters of information processed from operations of addition, subtraction, multiplication and division, which, to the best of our knowledge, has not been yet demonstrated in graph architectures.

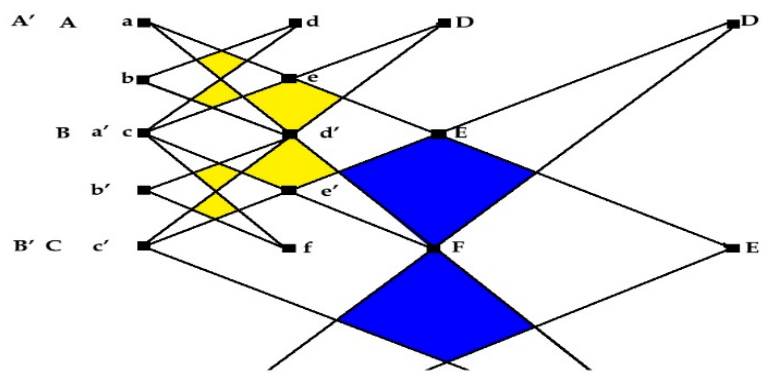

**Figure 8.** Searching at Various Query Scales of Cognitive Communities: self-similarity in GRAPHYP.

The scalability of GRAPHYP therefore provides a tool connecting approaches identified in different layers of the same information object; in SKG applications it could be particularly helpful to zoom inside the stages of the scientific domain or even to compare the evolutions of a given term used by different scientific domains. To the best of our knowledge, this property of self-similarity is not yet highlighted in current reviews of research on design of KGs (Ji et al., 2020). Such an architecture of self-similar graph could thus mobilize in an original way, the methods described in Wolfram for representation of Graph communities, like Random Graph or Community GraphPlot[23]. It could also give an additional solution to problems of Distributed Large-scale Natural Graph Factorization (Ahmed et al., 2013) and even provide a framework for large-scale graph decomposition and inference and give a research path on "automating the expansion of knowledge graphs" (Yoo & Jeong, 2020) in directions where bifurcations in sense building must be found, and therefore mainly in the applications of SKGs to Cognitive Communities.

The mapping of information retrieval routes thus becomes possible, like that of air or sea routes. Cooperative or connecting routes can be identified by GRAPHYP data structure. This last self-similarity property can help in drawing routes by proposing scaling and zooming functions from SKG GRAPHYP representations.

## 4. Discussion and Further works

---

## 4.1    Discussion

With GRAPHYP, we have designed an automated representation of the content consulted, exploiting what is known as "predecessor information". It required a non-trivial modeling of adversarial subnetworks through explainable paths of reasoning, to let the user choose, in a self-supervised attitude, a version of the knowledge best adjusted to his/her own hypotheses. As the analysis of cognitive structures is entering a new phase, driven by geometric graphs that are learning cognitive manifolds, GRAPHYP toolkit appears as an illustration at the intersection of those complementary evolutions: it is a graph designed not to represent information, but to model information representation. However, as noted in Section 1, SKGs still remain insufficiently effective in scholarly communication, and we are faced here with two challenges: the first is linked to a burgeoning scientific attention for this approach and the second is linked to the innovative direction of our approach, even if the first test is clearly successful. Both explain the care taken in our demonstrative precautions.

The first tests carried out and presented in Subsection 3.3 used query logs whose search terms themselves were not known. More complete tests exploiting a detailed semantic analysis of query logs are in preparation.

## 4.2    Further works

Further works will focus on the study of the modeling of SKGs with alternated relative symmetries in information representations in graphlets architectures, applied to the analysis of hypertext links of scientific results on the Web. Hypertexts could be approached as graphs whose nodes are texts and whose links are not random, in a context of an indirect communication mediated by a modification of the environment in the so-called « social stigmergy » (Marsh & Onof, 2008).

## 5. CONCLUSION

Two main conclusions emerge from this article.

One is that diversity matters in the SKG design. We have introduced here a new SKG that is consistent with the adversarial nature of the cognitive process in scholarly communication.

The second is that scientists shall derive unexplored benefits from representations of diversity. We develop an approach to "alternatives" in the results of search activities, as a new ancillary support to the discovery strategy, fully exploiting the potential of digital information architecture and its ability to address the design of diversity through exploitation of search history.

As other more general remarks, we propose three observations from the GRAPHYP experience.

The SKG is a structured interaction. Knowledge is not embedded in the graph, and the graph itself does not create knowledge. Knowledge graph is a mixed composite facility, the room of an interaction.

The diversity of Information is the fuel of rich interactions. "Heterosis", the sharing of diversity, remains a fruitful means of enrichment: GRAPHYP's final outputs thus combine the strengths of mutual information and of multiverse services.

The contextualized diversity in SKGs allow users and items to come closer. We are now starting to prepare a new stage of safe and genuine "contextualized networks", exploring all controversies, protected against fakes and rumors not by new walls, but by sharing between humans the complete representations of differences, reinforced by new skills and integrity standards, to provide scalable access to all homonymous and neighboring communities and routes, which could be expressed in any query.

## Supplementary files

A notebook describing the test procedure (subsection 3.3), is available as a supplementary file on Github: https://github.com/pbellot/graphyp.